\documentclass[runningheads]{llncs}
\usepackage[T1]{fontenc}
\usepackage{amsmath}
\usepackage{graphicx}
\usepackage{hyperref}
\begin{document}
\title{Investigating the Impact of Various Loss Functions
and Learnable Wiener Filter for Laparoscopic
Image Desmoking}
%
%
\author{Chengyu Yang\text{*} \and Chengjun Liu}
\authorrunning{Chengyu et al.} 
%
%
\institute{New Jersey Institute of Technology, Newark NJ 07103, USA,\\
\email{\{cy322\text{*}, chengjun.liu\}@njit.edu},
\\
WWW home page:\\
\url{https://chengyuyang-njit.github.io/}\\
\url{https://web.njit.edu/~cliu/}
}

\maketitle              
\begin{abstract}
To rigorously assess the effectiveness and necessity of individual components within the recently proposed ULW framework for laparoscopic image desmoking, this paper presents a comprehensive ablation study. The ULW approach combines a U-Net based backbone with a compound loss function that comprises mean squared error (MSE), structural similarity index (SSIM) loss, and perceptual loss. The framework also incorporates a differentiable, learnable Wiener filter module. In this study, each component is systematically ablated to evaluate its specific contribution to the overall performance of the whole framework. The analysis includes: (1) removal of the learnable Wiener filter, (2) selective use of individual loss terms from the composite loss function. All variants are benchmarked on a publicly available paired laparoscopic images dataset using quantitative metrics (SSIM, PSNR, MSE and CIEDE-2000) alongside qualitative visual comparisons.

\end{abstract}
\section{Introduction}
Laparoscopic procedures often suffer from degraded visibility due to the presence of surgical smoke, which can obscure critical anatomical structures and compromise both surgical precision and patient safety\cite{1}. Recent advancements in deep learning have led to significant improvements in image enhancement and restoration techniques aimed at addressing this issue\cite{2}.Among them, a novel framework referred to as ULW\cite{3} has been proposed for laparoscopic image desmoking. This method combines a U-Net\cite{4} based backbone with a compound loss function comprising mean squared error (MSE), structural similarity index (SSIM) loss\cite{5}, and perceptual loss\cite{6}. Additionally, this framework integrates a differentiable, learnable Wiener filter\cite{7} designed to further enhance output quality.

ULW has demonstrated the ability of produced desmoked images with both high visual clarity and strong performance across quantitative metrics, making it a promising solution for real-time surgical support. To rigorously evaluate the contribution of each component within the framework, this paper presents a comprehensive ablation study. Specifically, we examine the effects of: (1) removing the learnable Wiener filter and (2) selectively applying individual components of the composite loss function. Each model variant is evaluated on a publicly available paired laparoscopic image dataset using widely accepted quantitative metrics such as SSIM, PSNR, MSE and CIEDE-2000[8], as well as qualitative visual analysis. The findings provide insight into the relative importance of each design choice and validate the robustness of the overall approach.

\section{Overall Framework}
The ULW framework builds on a U-Net backbone, originally designed for biomedical image segmentation, to extract multi-scale features and retain spatial information through skip connections. In the context of laparoscopic image desmoking, this architecture enables the network to accurately identify and reconstruct clean anatomical structures from smoke-obscured inputs. By leveraging its encoder-decoder structure, U-Net effectively captures both global context and fine details essential for high-fidelity image restoration. The overall architecture is shown as below in Fig.\ref{fig1}. 
\begin{figure}
    \centering
    \includegraphics[width=0.95\linewidth]{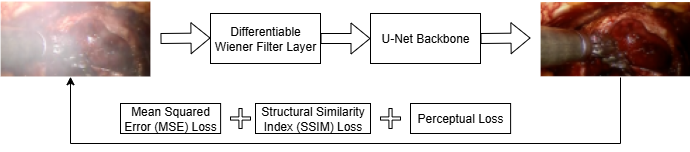}
    \caption{The framework of the ULW method}
    \label{fig1}
\end{figure}

\section{Components in details}
In this section, we introduce the details about separate components of the ULW method.

\subsection{Differentiable Wiener Filter Layer} The Wiener filter is a classical signal processing technique used for image denoising and deblurring. It operates by minimizing the mean squared error  MSE) between the estimated and true signals, assuming a linear degradation model and known signal/noise statistics. In the context of laparoscopic image desmoking, the degradation introduced by smoke, such as haze, blur, and reduced contrast, can be viewed as a noise-like process that distorts the underlying anatomical structures.

To model this process explicitly and guide the network with a physics-inspired prior, we introduce a differentiable, learnable Wiener filter layer into the ULW framework. Unlike fixed Wiener filters, which rely on hand-crafted assumptions, our implementation is trainable end-to-end, allowing it to learn optimal filter behavior directly from data. 

Let $x$ denote the input smoke-degraded image. The learnable Wiener filter estimates the clean image $\hat{x}$ using the following formulation:
\begin{equation}
    \hat{x} = s \odot \left( \frac{p}{p + \sigma^2 + \varepsilon} \right) \tag{1}
\end{equation}

$s = F(x)$ is the output of a convolutional filtering operation on $x$, initialized with a Gaussian kernel to simulate the smoothing nature of traditional Wiener filters.  $\odot$ represents element-wise multiplication. $p = s^2$ is the local signal power estimate, computed element-wise. $\sigma ^2$ is the learnable noise variance parameter, which enables the filter to adapt to different noise levels in various regions of the image.

This layer performs a spatially adaptive filtering, where each pixel is adjusted based on an estimate of the local signal-to-noise ratio (SNR). The term $\frac{p}{p + \sigma^2}$ acts as a soft gating mechanism: in regions where the signal is strong relative to the noise (high SNR), the gate value approaches 1, allowing the original signal to pass through. In heavily smoked regions, where the signal-to-noise ratio is low, the gating mechanism attenuates smoke-corrupted features, preventing them from degrading the restoration quality.

\subsection{Structural Similarity Index Measure (SSIM) Loss}
Traditional loss functions like Mean Squared Loss of $L1$ loss focus on pixel-wise differences between two images\cite{5}. While effective at enforcing numerical accuracy, these losses can lead to blurry or over smoothed outputs, especially in image restoration tasks like deblurring, denoising, or desmoking.

However, the human visual system is more sensitive to structures, contrast, and luminance rather than exact pixel matches\cite{9}. To better align with human perception, the Structural Similarity Index Measure (SSIM) was introduced.

In the context of laparoscopic image desmoking, where clear visualization of tissue structure and texture is crucial, using SSIM loss helps the model preserve fine anatomical details that might be lost under purely pixel-based losses.

Given two image patches $x$ and $y$, the SSIM index is defined as
\begin{equation}
    SSIM(x, y) = \frac{(2\mu_x \mu_y + C_1)(2\sigma_{xy} + C_2)}
{(\mu_x^2 + \mu_y^2 + C_1)(\sigma_x^2 + \sigma_y^2 + C_2)} \tag{2}
\end{equation}
where $\mu_x$ and $\mu_y$ are means of patches $x$ and $y$, $\sigma_x^2$ and $\sigma_y^2$ are variances, $\sigma_{xy}$ is the covariance between $x$ and $y$. $C_1$ and $C_2$ are small constants added to stabilize division. The SSIM value ranges from 0 to 1 where 1 means perfect structural similarity while 0 means no similarity. To use SSIM as a loss function, its complement is minimized as followed:
\begin{equation}
    L_{SSIM(x,y)} = 1 - SSIM(x, y) \tag{3}
\end{equation}

This loss encourages the network to maximize structural similarity between the predicted image and the ground-truth image.

In implementation, SSIM is usually computed over sliding windows (e.g. 11×11), and the total SSIM loss is averaged over the image.

\subsection{Perceptual Loss}
MSE or SSIM loss do not capture how humans perceive texture, structure, and fine-grained details. As a result, restored images may be numerically accurate but visually unconvincing. It’s a critical issue in laparoscopic desmoking, where clinicians rely on high-fidelity visuals to make surgical decisions\cite{10}.

To overcome this limitation, perceptual loss, which is also known as feature reconstruction loss, was introduced to encourage high-level similarity between the predicted and ground-truth images by comparing their deep feature representations rather than pixel values.

The idea is that if two images have similar deep features, they are likely to look similar to the human eye, even if their pixel values differ. Perceptual loss leverages a pretrained convolutional neural network (usually VGG-16 or VGG-19\cite{11} trained on ImageNet\cite{12}) to extract semantic features from both the predicted and target images. These features, captured in intermediate layers of the network, represent edges, textures, and object parts that are perceptually meaningful.

Let $x_{pred}$ be the predicted (desmoked) image, $x_{target}$ be the ground-truth smoke-free image. $ \phi_l$ be the feature map extracted from the l-th layer of a pretrained VGG network.
\begin{equation}
    L_{perceptual} = \left\| \phi_l(x_{pred}) - \phi_l(x_{target}) \right\|_2^2 \tag{4}
\end{equation}

\section{Ablation Study}

\begin{figure}
    \centering
    \includegraphics[width=0.95\linewidth]{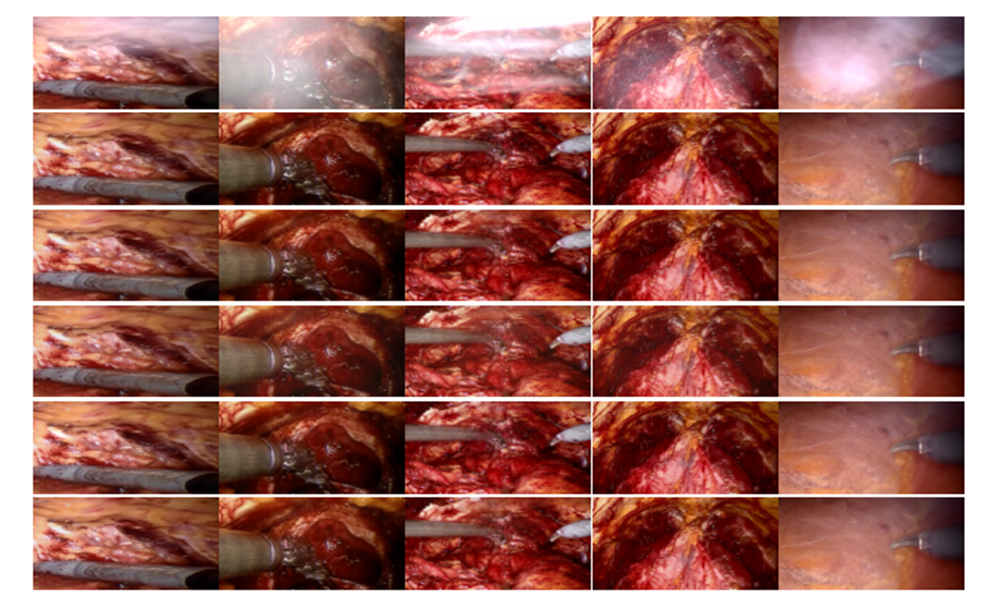}
    \caption{The visual presentation of the desmoking results produced by the ablated models. The first two rows show the paired laparoscopic images with and without smoke, respectively. The third row displays the results of the ULW method without the learnable Wiener filter layer, the fourth row reveals the results of the ULW method without the SSIM loss, the last but one row shows the results of the ULW method without the perceptual loss and the last row shows the results of the whole ULW framework.}
    \label{fig2}
\end{figure}

In this section, we conduct an ablation study by systematically removing individual components of the proposed framework described in the previous section. Each modified variant is evaluated both quantitatively—using SSIM, PSNR\cite{13}, MSE, and CIEDE-2000 metrics—and qualitatively through visual inspection. PSNR is the peak signal-to-noise ration which is the higher the better. It quantifies the ratio between the maximum possible signal value and the power of distortion (noise). CIEDE-2000 is a perceptual color difference metric that quantifies how different two colors appear to a human observer, based on the CIELAB color space\cite{14}. It improves upon older color metrics by incorporating hue rotation, chroma weighting, and interactive compensation terms. Lower values indicate higher color fidelity, with values below 2 generally considered perceptually indistinguishable. The quantitative results are shown in Table 1 and  the qualitative visual results are presented in Fig\ref{fig2}. 

\begin{table}[ht]
\centering
\caption{The quantitative metrics of the ablation study.}
\begin{tabular}{lcccc}
\hline
\textbf{Methods} & \textbf{SSIM} $\uparrow$ & \textbf{PSNR} $\uparrow$ & \textbf{MSE} $\downarrow$ & \textbf{CIEDI-2000} $\downarrow$ \\
\hline
w/o Learnable Wiener filter & 0.9909 & 33.9055 & \textbf{0.0007} & 1.8136 \\
w/o SSIM loss               & \textbf{0.9675} & 29.1446 & 0.0016 & 3.1500 \\
w/o Perceptual loss         & 0.9933 & 35.8490 & 0.0005 & 1.3511 \\
Full ULW                    & 0.9907 & 33.7061 & 0.0006 & 1.8159 \\
\hline
\end{tabular}
\label{tab:ablation}
\end{table}
\subsection{Ablation of Learnable Wiener Filter}
Although the inclusion of the learnable Wiener filter leads to slightly lower SSIM and PSNR, it improves MSE, indicating better pixel-wise accuracy. This trade-off suggests the filter effectively suppresses localized noise and haze caused by surgical smoke.

In addition, visual quality improves in low-contrast regions, such as tissue folds and smoke-heavy areas—details often overlooked by global metrics. Acting as a physics-inspired denoising prior, the Wiener filter enhances generalization and complements the U-Net’s data-driven learning.

Thus, despite minor metric variations, the Wiener filter is justified for its role in improving visual clarity and preserving fine details critical in surgical imaging.

\subsection{Ablation of SSIM Loss}
Removing the SSIM loss results in a substantial performance drop across all evaluation metrics, particularly the SSIM score itself, which decreases from 0.9907 in the full ULW model to 0.9675 without SSIM loss. This indicates a notable degradation in structural similarity, confirming the critical role of SSIM loss in preserving local anatomical features and fine tissue structures.

In addition, the PSNR drops by more than 4 dB, and CIEDE-2000 nearly doubles, reflecting both a loss of signal fidelity and a deterioration in color consistency. This highlights that SSIM loss not only improves structural alignment but also complements other loss components to enhance overall perceptual quality.

Qualitative results further support this finding: images generated without SSIM loss exhibit blurred edges, reduced contrast, and less distinct organ boundaries. In contrast, incorporating SSIM loss helps maintain sharp contours and anatomical clarity, which are essential in surgical contexts where accurate visual cues are vital.

Therefore, the inclusion of SSIM loss is justified both quantitatively and qualitatively. 

\subsection{Ablation of Perceptual Loss}
Although removing perceptual loss yields slightly better quantitative metrics (e.g., higher SSIM and PSNR), the visual outputs tend to be over-smoothed and lack fine-grained texture. Perceptual loss addresses this by comparing high-level features extracted from a pretrained network (e.g., VGG), encouraging the model to reconstruct images that are visually and semantically closer to the ground truth.
  
Perceptual quality matters in medical imaging because fine textures like blood vessels, sutures or tissue edges may not align perfectly pixel-wise, but must be perceptually faithful. Perceptual loss reduces blurring and improves texture sharpness. It also enhances clinical interpretability without sacrificing too much numerical accuracy.

Furthermore, it improves the model’s generalization by guiding the network to focus on meaningful features rather than overfitting to pixel-wise accuracy. In scenarios where clean ground truth is limited or imperfect, perceptual loss provides a robust supervisory signal.

Therefore, despite a slight trade-off in pixel-based metrics, the inclusion of perceptual loss is justified by its ability to enhance visual realism, textural fidelity, and practical utility of the desmoked images.

\section{Future Work}
Future work may explore extending the ULW framework to video-based desmoking by incorporating temporal consistency, which would enable real-time application and reduce flickering across frames. Additionally, evaluating the model’s effectiveness through clinical feedback from surgeons could offer valuable insight into the practical utility and visual clarity of desmoked images in real surgical settings.

\section{Conclusion}
In this study, we presented a comprehensive ablation analysis of the ULW framework for laparoscopic image desmoking, which integrates a U-Net backbone, a compound loss function (comprising MSE, SSIM loss, and perceptual loss), and a differentiable, learnable Wiener filter. Each component was systematically evaluated to understand its individual contribution to overall performance. The results show that the learnable Wiener filter enhances pixel-level reconstruction and improves visual clarity in low-contrast, smoke-affected regions, despite small changes in SSIM and PSNR. The SSIM loss proved critical for preserving anatomical structures, with its removal leading to significant drops across all metrics. Although omitting perceptual loss slightly improved some numerical scores, it resulted in over-smoothed outputs lacking texture and realism. Overall, the full ULW framework achieves a strong balance between quantitative accuracy and perceptual quality, confirming the necessity and effectiveness of each component in producing clinically interpretable desmoked laparoscopic images.
%
%
%
%

\end{document}